\documentclass[conference]{IEEEtran}
\usepackage{times}

\usepackage[numbers]{natbib}
\usepackage{multicol}
\usepackage[bookmarks=true]{hyperref}

\usepackage[ruled,vlined]{algorithm2e}

\usepackage{graphicx}
\usepackage{subcaption}
\usepackage{tikz}
\usepackage{cuted}
\usepackage[normalem]{ulem}
\usepackage{amsmath}
\usepackage{amssymb}  

\usepackage{url}
\usepackage[utf8]{inputenc} 
\usepackage[T1]{fontenc}    
\usepackage{booktabs}       
\usepackage{amsfonts}       
\usepackage{nicefrac}       
\usepackage{microtype}      
\usepackage{xcolor}         
\usepackage[table, dvipsnames]{xcolor}
\usepackage{indentfirst}
\usepackage{float}
\usepackage{dblfloatfix}
\usepackage{caption}
\usepackage[accsupp]{axessibility}
\usepackage{multirow}
\usepackage{wrapfig}
\usepackage{cleveref}
\usepackage{calc}

\let\titleold\title
\renewcommand{\title}[1]{\titleold{#1}\newcommand{\thetitle}{#1}}
\def\maketitlesupplementary
{
\newpage
   \twocolumn[
    \centering
    \Large
    \textbf{\thetitle}\\
    \vspace{0.5em}Supplementary Material \\
    \vspace{1.0em}
   ] 
}

\newcommand{\tocline}[2]{%
\noindent #1 \leaders\hbox{.\kern0.5pt}\hfill #2\par
}

\def\paper{DexImit}

\begin{document}

\newcommand{\red}[1]{\textcolor{red}{#1}}
\newcommand{\sizhe}[1]{\red{(Sizhe: {#1})}}
\newcommand{\lnxu}[1]{\textbf{\color{purple}#1}}
\newcommand{\TODO}[1]{\textbf{\color{red}[TODO: #1]}}

\definecolor{myOrange}{RGB}{230,126,34}
\definecolor{myGreen}{RGB}{96,150,90}    
\definecolor{myPurple}{RGB}{128,96,170}  
\definecolor{myBlue}{RGB}{52,115,186}    

\definecolor{myBrown}{RGB}{235,195,90}

\newcommand{\parta}[1]{\textcolor{myOrange}{#1}}
\newcommand{\partb}[1]{\textcolor{myGreen}{#1}}
\newcommand{\partc}[1]{\textcolor{myPurple}{#1}}
\newcommand{\partd}[1]{\textcolor{myBlue}{#1}}

\def\best{\cellcolor{myOrange!40}}
\def\second{\cellcolor{yellow!30}}

\title{DexImit: Learning Bimanual Dexterous Manipulation from Monocular Human Videos}

\author{
\textbf{Juncheng Mu}$^{1,2,*}$,
\textbf{Sizhe Yang}$^{1,3,*}$,
\textbf{Yiming Bao}$^{2}$,
\textbf{Hojin Bae}$^{2}$,
\textbf{Tianming Wei}$^{2}$, \\
\textbf{Linning Xu}$^{1}$,
\textbf{Boyi Li}$^{4,\dagger}$,
\textbf{Huazhe Xu}$^{2,\dagger}$,
\textbf{Jiangmiao Pang}$^{1,\dagger}$ \\
$^{1}$Shanghai AI Laboratory ,
$^{2}$Tsinghua University,
$^{3}$The Chinese University of Hong Kong,
$^{4}$NVIDIA \\
\textcolor{gray}{$^{*}$Equal contribution} \qquad
\textcolor{gray}{$^{\dagger}$Corresponding author} \\
Project Page:\quad \href{https://mujc2021.github.io/deximit/}{\textcolor[HTML]{D81B60}{https://mujc2021.github.io/deximit/}}
}



%

\maketitle
\IEEEpeerreviewmaketitle
\begin{strip}
\centering
\vspace{-8mm}
\includegraphics[width=\textwidth]{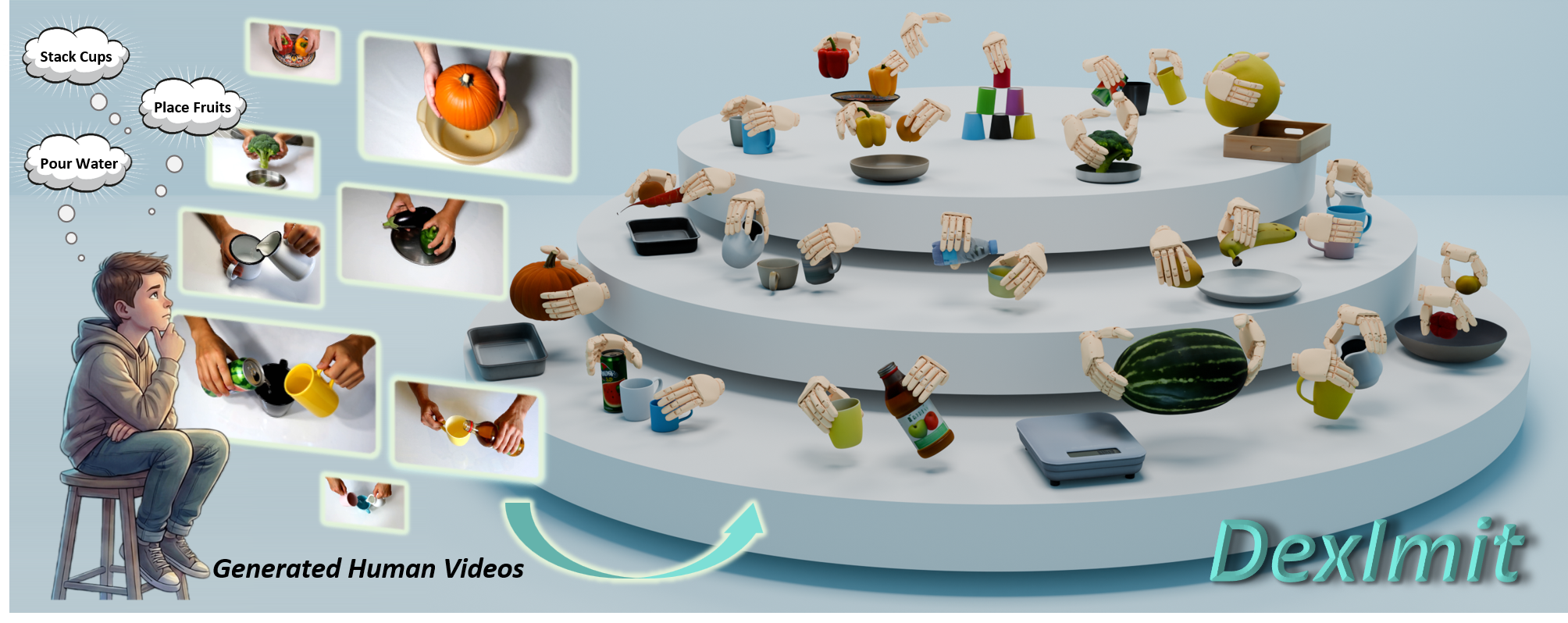}
\vspace{-3mm}
\captionof{figure}{
We introduce \paper, a framework for learning dexterous manipulation directly from videos.
\paper\ leverages generated or in-the-wild videos to synthesize physically plausible demonstrations, including challenging tool-using, long-horizon, and fine-grained tasks.
The gallery highlights the breadth of manipulation tasks generated by \paper.
}
\label{fig:teaser}
\end{strip}

\begin{abstract}
Data scarcity fundamentally limits the generalization of bimanual dexterous manipulation, as real-world data collection for dexterous hands is expensive and labor-intensive.
Human manipulation videos, as a direct carrier of manipulation knowledge, offer significant potential for scaling up robot learning.
However, the substantial embodiment gap between human hands and robotic dexterous hands makes direct pretraining from human videos extremely challenging.
To bridge this gap and unleash the potential of large-scale human manipulation video data, we propose \paper, an automated framework that converts monocular human manipulation videos into physically plausible robot data, without any additional information. 
\paper\ employs a four-stage generation pipeline:
(1) reconstructing hand-object interactions from arbitrary viewpoints with near-metric scale;
(2) performing subtask decomposition and bimanual scheduling;
(3) synthesizing robot trajectories consistent with the demonstrated interactions;
(4) comprehensive data augmentation for zero-shot real-world deployment.
Building on these designs, \paper\ can generate large-scale robot data based on human videos, either from the Internet or video generation models.
\paper\ is capable of handling diverse manipulation tasks, including tool use (e.g., cutting an apple), long-horizon tasks (e.g., making a beverage), and fine-grained manipulations (e.g., stacking cups).

\end{abstract}

\section{Introduction}
\label{sec:intro}

Data scarcity remains a critical challenge in dexterous manipulation.
Robotic hands are highly articulated and versatile end-effectors, capable of performing a wide range of real-world tasks, including contact-rich and interaction-intensive operations.
However, due to the difficulty of teleoperation and the high cost of hardware, collecting large-scale datasets for bimanual dexterous manipulation is significantly more challenging than for simple jaw-grippers \cite{pi0, pi05, gen0}.

Compared to real-world robotic data, human manipulation videos are readily available at a larger scale and cover a substantially broader range of task categories.
Furthermore, the emergence of video generation models \cite{wan2.2, sora2, pixverse_v5, seedance, veo3} enables scalable generation of human manipulation videos from text prompts.
Human videos inherently encode high-level task concepts while simultaneously capturing low-level manipulation actions, offering a promising avenue for scaling dexterous manipulation.

However, learning from human videos poses significant challenges.
The most straightforward approach is to treat the human hand as a heterogeneous embodiment and use it directly for pretraining \cite{pi_human, scalable_human_pretrain, egovla, beingh0, h-rdt}.
These methods suffer from a severe embodiment gap, as discrepancies in visual observations and action spaces substantially constrain cross-embodiment learning.
Another line of work reconstructs 3D hand-object keypoint flows \cite{novaflow, dream2flow, avdc, gen2act_variant} or object trajectories \cite{physworld, dexman, hermes, robowheel, rigvid} from videos, then reproduces the demonstrations using motion planning \cite{motionplan, curobo} or reinforcement learning (RL). This paradigm effectively eliminates the embodiment gap, but most of them rely on absolute depth information, while others \cite{dexman, physworld} require strict reconstruction accuracy to avoid RL training failures—fundamentally limiting scalability.
Moreover, existing methods struggle with challenging scenarios involving fast motions, occlusions, or complex interactions.

To address these challenges, we propose a four-stage framework: (1) depth-free \parta{\textbf{reconstruction}} with near-metric scale; (2) action-centric \partb{\textbf{scheduling}} for long-horizon bimanual coordination; (3) force-closure-based \partc{\textbf{action generation}} for robust manipulation and (4) comprehensive data \partd{\textbf{augmentation}} to cover complex real-world environments.
Through these designs, \paper\ can operate without additional depth or camera information, offering a scalable solution to alleviate the scarcity of bimanual dexterous manipulation data in the real world.
Specifically, our method first reconstructs hand-object trajectories from monocular human videos captured from arbitrary viewpoints, mapping them into a shared world coordinate system with near-metric scale.
We then perform video understanding and subtask decomposition, and introduce an \textit{Action-Centric Task Scheduling Algorithm} to enable dynamic bimanual coordination.
Finally, we synthesize grasps based on force-closure constraints and reconstructed hand poses, and reproduce the demonstrated interactions through motion planning.
To support zero-shot deployment in complex real-world environments, we design a comprehensive data augmentation pipeline.
It includes randomization of object pose and scale, as well as augmentation of camera pose and visual observations to ensure robust real-world generalization.

Building upon these designs, \paper\ is capable of generating bimanual dexterous manipulation data across diverse scenarios, including tasks involving complex physical interaction, unleashing the potential of large-scale human manipulation videos for robot learning.
Extensive experimental results further demonstrate that policies trained on the resulting data generalize to real-world deployment in a zero-shot manner.

In summary, our contributions are threefold:
\begin{itemize}
\item An automated data generation pipeline for bimanual dexterous manipulation, synthesizing physically plausible data covering a broad range of tasks directly from videos.
\item A comprehensive data augmentation system, including object pose and scale, as well as camera pose and visual observation, facilitating zero-shot deployment of policies on real robots without any real-world data.
\item Extensive experiments demonstrate that \paper\ can generate high-fidelity robot data for diverse tasks, including long-horizon, tool-using, and fine-grained manipulation, demonstrating its effectiveness in alleviating the longstanding data scarcity problem in dexterous manipulation.
\end{itemize}
\section{Related Works}
\label{sec:related_work}

\subsection{Learning from Videos}
With the emergence of large-scale human manipulation datasets \cite{dataset1,dataset2,dataset3,dataset4,dataset5,dataset6, motiontrans} and video generation models \cite{wan2.2, sora2, pixverse_v5, seedance, veo3, LVP}, an increasing number of studies have focused on learning transferable dexterous manipulation skills from human videos. A straightforward method is to treat the human hand as an end-effector and incorporate it directly into the policy pretraining \cite{pi_human, scalable_human_pretrain, egovla, beingh0, h-rdt, motiontrans}. However, such methods face fundamental limitations due to substantial visual and action embodiment gaps.
Other methods leverage human videos to train world models for planning; however, world models tailored specifically for manipulation remain relatively underexplored \cite{vpp, cosmos, LVP}. A separate line of work reconstructs 3D trajectories directly from videos and uses the recovered motions to synthesize robot manipulation data \cite{dexman, physworld, robowheel}, effectively bridging the embodiment gap.
Building on this paradigm, \paper\ effectively mitigates compounding errors of the data generation, enabling the synthesis of challenging long-horizon tasks involving complex physical interactions.

\subsection{Monocular Reconstruction}
To reconstruct hand-object interactions from videos without access to additional information (e.g., camera pose, intrinsic parameters, or depth), we must rely on 4D reconstruction from monocular observations.
VGGT \cite{vggt} and several recent methods \cite{mapanything, pi3, da3, spatialtrackerv2} perform feed-forward estimation of depth and camera parameters, enabling per-frame point cloud reconstruction, which can then be combined with hand-object segmentation \cite{sam, sam2, grounded_sam} and point cloud registration \cite{ICP, geotransformer, colorpcr} to achieve tracking.
End-to-end tracking methods \cite{pomato, st4rtrack, cotracker3, traceanything, spatialtrackerv2} instead directly estimate foreground point cloud motion.
However, these approaches are often limited in accuracy, producing point clouds of insufficient quality for reliable manipulation.
With recent advances in image-to-3D generation \cite{trellis, hunyuan3d, sam3d} and hand pose estimation \cite{hammer, wilor}, single-image 3D mesh generation has become sufficiently accurate for downstream manipulation tasks.
As a result, hand-object interactions can be reconstructed based on the generated meshes.
Finally, 6D pose estimation methods \cite{foundpose, any6d, oneposeviagen} are used to estimate object poses at each video frame.

\begin{figure*}[t!]
\vspace{-7mm}
\centering
\includegraphics[width=0.98\textwidth]{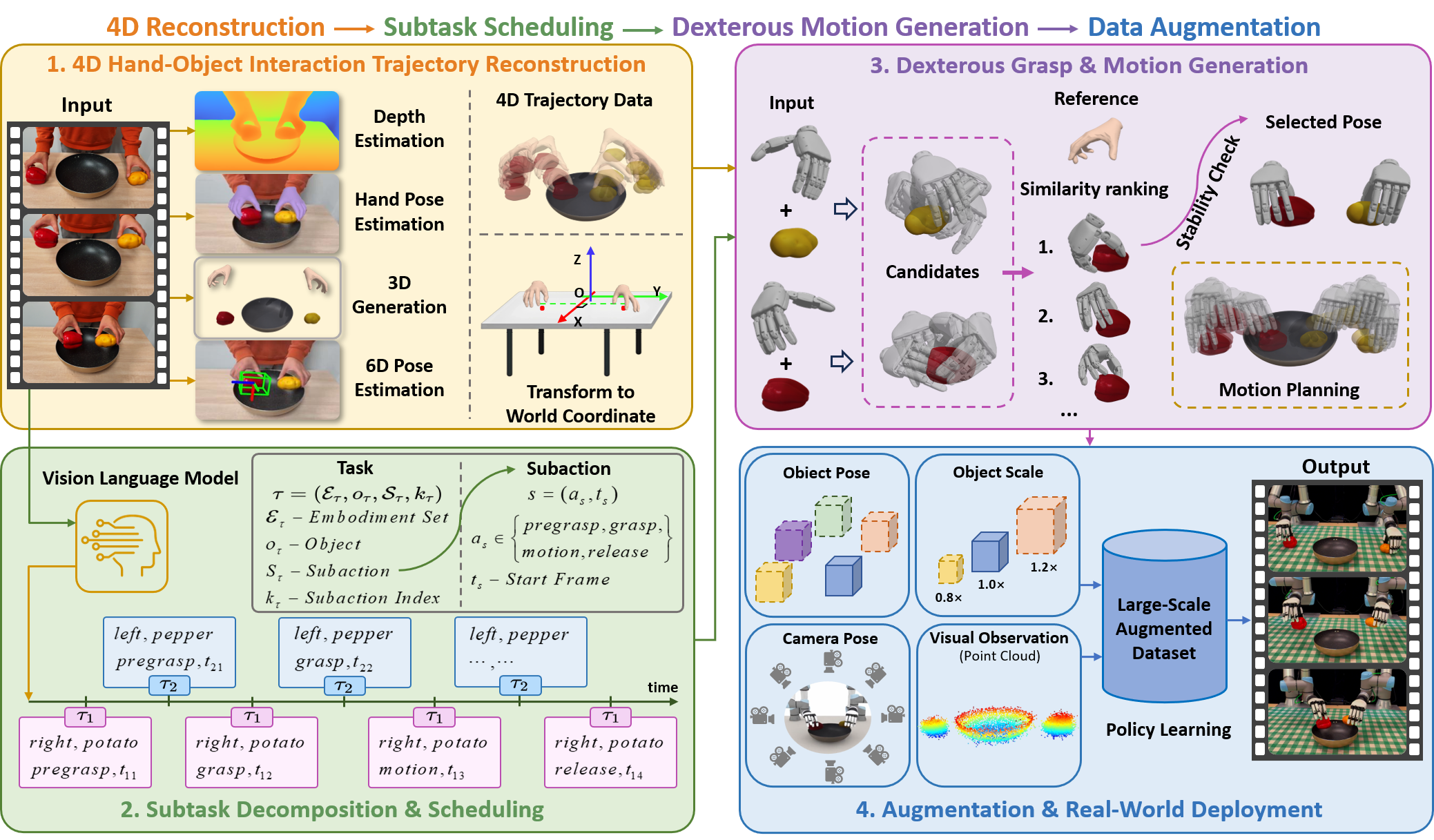}
\caption{
We adopt a four-stage paradigm: \parta{\textbf{Reconstruction}}-\partb{\textbf{Scheduling}}-\partc{\textbf{Action}}-\partd{\textbf{Augmentation}}.
(1) Reconstruct 4D hand-object interactions and transform them to a unified world frame.
(2) Decompose the manipulation process into subtasks and schedule bimanual actions for long-horizon tasks using an \textit{Action-Centric Scheduling Algorithm}.
(3) Generate robot trajectories via grasp synthesis and motion planning.
(4) Augment the resulting source data comprehensively to enable robust policy learning.
}
\vspace{-6mm}
\label{fig:pipeline}
\end{figure*}

\subsection{Robot Data Generation from Reconstructed Reference}
Given the reconstructed hand-object poses, our goal is to generate robot manipulation data from these trajectories. Prior approaches typically rely on reinforcement learning \cite{videorl1,videorl2,videorl3,videorl4,videorl5,videorl6,videorl7,videorl8,videorl9,videorl10,dexman,physworld} or motion retargeting \cite{rigvid, avdc, gen2act_variant, robowheel, igen, rola} to reproduce hand-object interactions. However, these methods are sensitive to object scale variations and trajectory noise, suffer from substantial sim-to-real gaps, and remain inadequate for long-horizon bimanual dexterous manipulation.
In contrast, \paper\ introduces a multi-arm long-horizon task scheduling algorithm and synthesizes structured, robust actions based on MANO-prompts and force-closure constraints, enabling reliable generation of physically plausible manipulation data.
\section{Method}
\label{sec:method}

We propose \paper, a framework that directly generates bimanual dexterous manipulation data from human videos.
\paper\ employs a four-stage data generation pipeline:
(1) 4D trajectory \parta{\textbf{reconstruction}} of hand-object interactions (\Cref{method:4d reconstruction});
(2) subtask decomposition and bimanual \partb{\textbf{scheduling}} (\Cref{method:subtask decomposition});
(3) structured \partc{\textbf{action generation}} to produce robot manipulation trajectories (\Cref{method:src generation});
and (4) comprehensive data \partd{\textbf{augmentation}} (\Cref{method:augmentation}).
The overall pipeline is illustrated in \Cref{fig:pipeline}.
This design enables \paper\ to generate dexterous manipulation data at scale and supports zero-shot real-world deployment.

\vspace{-5mm}
\parta{
\subsection{\textbf{Reconstruction of 4D Hand-Object Interactions}}
\label{method:4d reconstruction}
}

The reconstruction of hand-object interactions involves the following steps:
(1) Video processing and task understanding using a Vision-Language Model; (2) Frame-by-frame semantic segmentation to isolate objects involved in manipulation; (3) Object generation and hand pose estimation for each frame; (4) 6D pose estimation for both the object and hand to obtain precise trajectories. (5) Coordinate transformation from camera coordinates to a shared world frame.

\paragraph{\textbf{Video Process}}
Given a video $ V = \{I_i\}_{i=0}^{K} $ with a frame rate \( f \), we sample it to a constant frame rate \( f_t \) to obtain the target video:
$V_t = \left\{ I_{\left\lfloor \frac{i \cdot f}{f_t} \right\rfloor}\right\}_{i=0}^{K_t},$
where $K_t = \lfloor \frac{K \cdot f_t}{f}\rfloor - 1$ is the last index of sampled frames.
Next, we employ Qwen3-VL \cite{Qwen3-VL} to perform video understanding on the \( V_t \). The model is tasked with identifying the set of objects \( S_o = \{ o_i \mid i=0,1, \dots, N_o\} \) involved in the manipulation process.

\paragraph{\textbf{Segmentation}}

The data generation pipeline requires three types of masks:
\textit{Object mask} $m_o = \{m_{o_i}\}_{i=0}^{N_o}$ for 3D generation and 6D pose estimation;
\textit{Hand mask} $m_h = \{m_{h_i}\mid i=0,1\}$ for hand trajectory estimation, where $h_0$ denotes the left hand and $h_1$ denotes the right hand;
\textit{Table mask} $m_t$ for determining the world coordinate system.
Specifically, we use Grounded Sam2 \cite{grounded_sam} to perform frame-by-frame segmentation to generate these three masks.

\paragraph{\textbf{Objects and Hands Reconstruction}}
In order to reconstruct hand-object interactions at near-metric scale, we first use the depth estimation method SpatialTracker v2 \cite{spatialtrackerv2} to estimate the unscaled depth $D=\{D_i\}_{i=0}^{K_t}$ for each frame of the video.
Since the input RGB video $V_t$ does not contain any depth information, an initial scale estimation step is necessary. Drawing inspiration from \cite{dexman}, the limited variance in human hand sizes provides a reliable prior to approximate metric scale.
Building on this key insight, we use the human hand to estimate a scale factor for $D$.

Take left hand $h_0$ as an example, with the first frame hand mask $m_{h_0}^0$, we extract the first-frame hand point cloud $\mathcal{P}_{h_0}^0$. Next, we use Wilor \cite{wilor} to estimate hand mesh $\mathcal{M}_{h_0}^0$. While this provides the correct orientation of the hand in the camera coordinate system, it lacks position information. To jointly estimate both the hand pose and metric scale, we proceed with the following \textbf{align-render-align} steps:
we first translate the center of $\mathcal{M}_{h_0}^0$ to align it with the center of $\mathcal{P}_{h_0}^0$ for coarse alignment. Since $\mathcal{M}_{h_0}^0$ is a full hand reconstruction, while $\mathcal{P}_{h_0}^0$ only contains visible parts of the camera, the coarse alignment introduces a depth gap.
To address this, we cast parallel light rays in the camera coordinate system to calculate the occlusion-free mesh $\mathcal{\hat{M}}_{h_0}^0$. These vertices are then translated to the center of $\mathcal{P}_{h_0}^0$.
We then calculate the scale factor $s = \frac{PCA(\mathcal{\hat{M}}_{h_0}^0)}{PCA(\mathcal{P}_{h_0}^0)}$, where $PCA$ \cite{pca} computes the length of the principal component axis of the point cloud. Then, \( s \) is applied to \( D \), thereby obtaining the metric scale depth \( \hat{D} = \{ \hat{D}_i\}_{i=0}^{K_t} \).

We then use SAM3D \cite{sam3d} to perform image-to-3D generation, which can generate objects with approximate initial pose and scale. To achieve accurate scale alignment, we repeat the previously described \textbf{align-render-align} procedure, but this time the estimated scale factor is applied to the object, achieving $\mathcal{M}_o=\{\mathcal{M}_{o_i}\}_{i=0}^{N_o}$ aligned with $\hat{D}$.
For the hand, we directly use the per-frame poses $\mathcal{M}_h$ estimated by Wilor \cite{wilor} without applying any additional scale adjustment.

\paragraph{\textbf{6D Pose Estimation}}
After obtaining $\mathcal{M}_h, \mathcal{M}_o$ at near-metric scale, we perform 6D pose estimation to reconstruct the full manipulation trajectories.
For $\mathcal{M}_o$, we adopt a tracking variant \cite{FoundationPosePlusPlus} of FoundationPose \cite{foundpose}, which effectively improves temporal continuity of pose estimation across frames.
For $\mathcal{M}_h$, since they already have accurate orientation, we only need to estimate the translation component. We apply the previously described \textbf{align-render-align} procedure to recover continuous hand poses over time.
At this point, we have completed the full reconstruction of both hand trajectories $\{p_h^t\}_{t=0}^{K_t}$ and object trajectories $\{p_o^t\}_{t=0}^{K_t}$.

\paragraph{\textbf{World Coordinate Transformation}}
Since the camera viewpoints of the videos are arbitrary, 
we map all reconstructed trajectories into a fixed world coordinate system (see \Cref{fig:pipeline}).
Our key insight is that the transformation from the camera frame to the world frame can be uniquely determined by specifying the world-frame origin and the orientations of the $x$- and $z$-axes in the camera frame, after which the $y$-axis is uniquely defined by the right-handed coordinate system constraint.

We denote the camera coordinate frame as $\mathcal{F}_c$ and the world coordinate frame as $\mathcal{F}_w$.
The transformation from $\mathcal{F}_c$ to $\mathcal{F}_w$ is represented by a homogeneous transformation
$\mathbf{T}_{c \rightarrow w} \in SE(3)$, which can be decomposed as:
\begin{equation}
\mathbf{T}_{c \rightarrow w} =
\begin{bmatrix}
\mathbf{R}_{c \rightarrow w} & \mathbf{t}_{c \rightarrow w} \\
\mathbf{0}^\top & 1
\end{bmatrix},
\end{equation}
where $\mathbf{R}_{c \rightarrow w} \in SO(3)$ and $\mathbf{t}_{c \rightarrow w} \in \mathbb{R}^3$ denote the rotation
and translation, respectively.

\noindent
\textbf{World axis estimation}.
We compute the world coordinate axes in the camera frame as follows.

(1) $z$-axis.
Given the table segmentation mask $m_t$, we extract the table point cloud and estimate its surface normal.
The world $z$-axis is defined as the normalized normal vector of the table plane: $\mathbf{z}_w = \frac{\mathbf{n}_t}{\|\mathbf{n}_t\|}.$

(2) $x$-axis.
We compute the first-frame perpendicular bisector direction of $p_{h_0}^0$ and $p_{h_1}^0$.
This direction is projected onto the plane orthogonal to $\mathbf{z}_w$ to ensure orthogonality:
\begin{equation}
\tilde{\mathbf{x}}_w = \mathbf{d}_h - (\mathbf{d}_h^\top \mathbf{z}_w)\mathbf{z}_w, \quad
\mathbf{x}_w = \frac{\tilde{\mathbf{x}}_w}{\|\tilde{\mathbf{x}}_w\|},
\end{equation}
where $\mathbf{d}_h$ denotes the hand-derived direction vector.

(3) $y$-axis determination.
The world $y$-axis is uniquely determined by the right-handed coordinate system constraint.

(4) Rotation matrix construction.
The rotation matrix from the camera frame to the world frame is thus given by
\begin{equation}
\mathbf{R}_{c \rightarrow w} =
\begin{bmatrix}
\mathbf{x}_w & \mathbf{y}_w & \mathbf{z}_w
\end{bmatrix}.
\end{equation}

\noindent
\textbf{Origin estimation}.
We define the world-frame origin based on the object poses in the first video frame.
Specifically, we compute the center of the axis-aligned bounding box enclosing all manipulated objects and translate it to a predefined manipulation region in the robot workspace ($x=0.6$ in the world frame), yielding the translation vector $\mathbf{t}_{c \rightarrow w}$.

With the above procedure, the transformation $\mathbf{T}_{c \rightarrow w}$ is fully determined,
allowing us to map arbitrary-view human videos into a unified world coordinate system.

\vspace{-4mm}
\partb{
\subsection{\textbf{Subtask Decomposition and Task Scheduling}}
\label{method:subtask decomposition}
}
\paper\ places no restrictions on the temporal horizon of the input video, the degree of bimanual concurrency or asynchrony, or the types of manipulation actions involved—ranging from unimanual actions (e.g., grasping), to cooperative bimanual actions (e.g., bimanual grasping), and fully concurrent bimanual actions (e.g., pouring).
To enable scheduling under arbitrary horizons and action combinations, we introduce a dedicated \emph{Action-Centric Scheduling} algorithm.

\noindent
\textbf{Task Representation.}
We define two structures:
\begin{itemize}
    \item A \textbf{Task} $\tau$ is defined as $\tau = \big( \mathcal{E}_\tau,\; o_\tau,\; \mathcal{S}_\tau,\; k_\tau \big)$,
    where $\mathcal{E}_\tau \subseteq \{1, \dots, N\}$ denotes the \emph{embodiment set} involved in the task, $o_\tau$ is the associated object, $\mathcal{S}_\tau$ is the ordered list of subactions, and $k_\tau$ is the current subaction index.

    \item A \textbf{Subaction} $s \in \mathcal{S}_\tau$ is defined as $s = \big( a_s,\; t_s \big)$,
    where $t_s$ denotes the start frame, and the action type
    $a_s \in \{\texttt{pregrasp}, \texttt{grasp}, \texttt{motion}, \texttt{release}\}$
\end{itemize}

\noindent
\textbf{Task Annotation.}
To annotate the above task structures from input videos, we employ Qwen3-VL \cite{Qwen3-VL} for video understanding, subtask decomposition, and structured labeling. For long-horizon tasks, we optionally incorporate manual annotations to further improve labeling accuracy.

{
\begin{algorithm}
\footnotesize

\caption{Action-Centric Scheduling}
\label{alg:embodiment_scheduling}
\KwIn{Number of embodiments $N$, horizon $T$, tasks $\mathcal{L}_t$}
\KwOut{Action queues $\{\mathcal{A}_i\}_{i=1}^{N}$}

Sort $\mathcal{L}_t$ by start time; initialize priority queue $\mathcal{Q}$ with first $N$ tasks; initialize all-zero action set $\mathcal{A}_i \in \mathbb{R}^{T}$ for each embodiment $i \in \{1,\dots,N\}$;

\For{$t = 1$ \KwTo $T$}{
    \ForEach{task $\tau \in \mathcal{Q}$}{
        \While{$|\mathcal{Q}| < N$ \textbf{and} $\mathcal{L}_t \neq \emptyset$}{
            Pop next task from $\mathcal{L}_t$ and push into $\mathcal{Q}$
        }
        Let $s = \mathcal{S}_\tau[k_\tau]$\;
        \If{$t = t_s$}{
            \Switch{$a_s$}{
                \Case{\texttt{pregrasp}}{Grasp generation candidates and motion planning for $\mathcal{E}_\tau$}
                \Case{\texttt{motion} or \texttt{grasp}}{Motion planning for $\mathcal{E}_\tau$}
                \Case{\texttt{release}}{Reset  joints for $\mathcal{E}_\tau$}
            }
            Assign trajectories to $\mathcal{A}_i$; $k_\tau \leftarrow k_\tau+1$\;
            \If{$k_\tau > |\mathcal{S}_\tau|$}{Remove $\tau$ from $\mathcal{Q}$}
        }
    }
    \For{$i = 1$ \KwTo $N$}{
        Execute $\mathcal{A}_i[t]$ if not empty
    }
}
\end{algorithm}
}

\noindent
\textbf{Action-Centric Scheduling.}
Based on the extracted task structures, we design an \textbf{Action-Centric Scheduling Algorithm} to perform subtask scheduling across arbitrary numbers of embodiments, horizons and action combinations. The overall procedure is summarized in \Cref{alg:embodiment_scheduling}.

\vspace{-4mm}
\partc{
\subsection{\textbf{Source Data Generation}}
\label{method:src generation}
}
After reconstructing the hand-object trajectories and scheduling actions via Algorithm~\ref{alg:embodiment_scheduling}, 
we perform low-level action generation through force-closure-based grasp synthesis and key-frame-based motion planning.

\paragraph{\textbf{Grasp Synthesis}}
Suppose that at time step $t$, embodiments $\mathcal{E}_\tau \subseteq \{h_0, h_1\}$ are required to generate a grasp on object $o_i$.
We adopt a candidate generation-and-selection strategy to synthesize such grasps.
Grasp synthesis is carried out in two stages: initialization and subsequent optimization to ensure physical feasibility.
During initialization, we first compute the convex hull of the object mesh. For unimanual grasps, a single contact point is sampled on the hull; for bimanual grasps, two contact points are sampled on opposite sides with respect to the object center. The hands are then initialized along the corresponding surface normals, with palms oriented toward the object, yielding a diverse set of feasible starting poses.

Following BODex~\citep{bodex}, we formulate the problem using an object mesh $\mathcal{M}_o$ with center of mass $\mathbf{m}$, hands $h \in \{h_0,h_1\}$, and a strategy-specific active contact set $\mathcal{C}$.
The decision variables are the bimanual pose $\mathbf{g} = \{ (\mathbf{t}_h, \mathbf{R}_h, \mathbf{q}_h) \}$ and contact forces $\{ \mathbf{f}_{\mathbf{c}} \}$, where $\mathbf{t}_h,\mathbf{R}_h$ and $\mathbf{q}_h$ denote the translation, rotation and joint angles of hand $h$. Let $\mathbf{p}_{\mathbf{c}}$ and $\mathbf{O}_{\mathbf{c}}$ represent the forward kinematics-derived contact position and orientation, defining the grasp map $\mathbf{G}_{\mathbf{c}} = [\mathbf{I}; (\mathbf{p}_{\mathbf{c}} - \mathbf{m})_{\times}]\mathbf{O}_{\mathbf{c}}$.
Given target wrenches $\{\mathbf{w}_j\}$ and scaling $\lambda$, and weights $\kappa_\bullet$, the grasp synthesis minimizes:
\begin{equation}
\begin{aligned}
&\min_{\mathbf{g},\,\{\mathbf{f}_{\mathbf{c}}\}}\quad
 \kappa_{w} \sum_{j=1}^J \Big\| \lambda \mathbf{w}_j - \sum_{{\mathbf{c}}\in\mathcal{C}} \mathbf{G}_{\mathbf{c}}(\mathbf{g}) \mathbf{f}_{\mathbf{c}} \Big\|_2^2 \\
& \quad + \kappa_{\text{con}} \sum_{{\mathbf{c}}\in\mathcal{C}} \psi(d_M(\mathbf{p}_{\mathbf{c}})) \ + \kappa_{\text{coll}} \,\Phi_{M}(\mathbf{g}) + \kappa_{\text{hh}} \,\Phi_{\text{hh}}(\mathbf{g}).
\end{aligned}
\label{eq:main_obj}
\end{equation}
The terms $\psi(d_M(\cdot))$, $\Phi_{M}$, and $\Phi_{\text{hh}}$ penalize contact distance, hand-object collision, and hand-hand penetration, respectively.
Solving Eq.~\eqref{eq:main_obj} yields a set of physically feasible and collision-free grasp candidates 
\(\mathcal{G}_{o_i}\) for object \(o_i\).

Then we rank the grasp candidates in $\mathcal{G}_{o_i}$ according to their distance to the reconstructed human hand pose $p_{\mathcal{E}_\tau}^t$ at time $t$, favoring candidates that are more consistent with the demonstrated human hand behaviors.
\begin{equation}
\mathcal{G}_{o_i}^{\text{sorted}} = \text{sort}\Big(\{g_j\}, \; d(g_j, p_{\mathcal{E}_\tau}^t)\Big),
\end{equation}
where $d(\cdot,\cdot)$ denotes a distance metric between hand pose and grasp candidate.
We then sequentially evaluate the stability of each candidate $g_j \in \mathcal{G}_{o_i}^{\text{sorted}}$ until finding a candidate that satisfies the stability criterion, denoted as the final grasp $g^*$; details are provided in Appendix \hyperref[sec:supp method]{A}
.

\paragraph{\textbf{Motion Generation}}
\paper\ performs motion planning based on the key-frame object poses.
Let $p_{o_i}^t$ and $p_{o_i}^{t'}$ denote the object poses at the current time $t$ and target time $t'$, respectively. 
The relative transformation is computed as
\begin{equation}
\mathbf{T}_{o_i}^{t\rightarrow t'} = \big(p_{o_i}^t\big)^{-1} \, p_{o_i}^{t'}.
\end{equation}
We treat the hand and object as a single rigid body after grasping.
Applying this transformation to the end-effector pose of the selected embodiment(s) at time $t$, denoted as $p_{\text{ee},\mathcal{E}_\tau}^t$, yields the target end-effector pose at time $t'$:
\begin{equation}
p_{\text{ee},\mathcal{E}_\tau}^{t'} = \mathbf{T}_{o_i}^{t\rightarrow t'} \, p_{\text{ee},\mathcal{E}_\tau}^t.
\end{equation}
This provides the terminal configuration for motion planning of the embodiment(s) to achieve the desired object manipulation.

\vspace{-4mm}
\partd{
\subsection{\textbf{Data Augmentation}}
\label{method:augmentation}
}

To generate large-scale data for policy learning, we apply \textbf{comprehensive data augmentation} to the source trajectories. Specifically, we perform four types of augmentation:

\textbf{Object pose:} Following \cite{demogen, robosplat}, we randomize the \emph{position} and \emph{translation} of objects to enable spatial generalization.

\textbf{Object scale:} To generalize across diverse object scales and enable zero-shot deployment in the real world, we augment the reconstructed near-metric object scales. The source data is deemed as a 1.0 scale; in practice, we apply scale factors of $[0.8, 1.2]$.
We empirically observe that, under our dp3-based policy learning setting, regenerating grasps and motions for each object scale leads to inconsistent supervision across demonstrations, which destabilizes training and slows convergence (see \Cref{sec:exp:sim2real}).
Therefore, we retain the original source grasps and motions and only adjust finger articulation to accommodate the augmented scales.

\textbf{Camera pose:} To enable viewpoint generalization, we randomize both the \emph{orientation} and \emph{position} of the camera.

\textbf{Observation:} We use 3D point clouds as the observation. To simulate the variability of real-world depth sensors, we apply augmentation to the object point clouds by randomly removing 30\% of the points and adding noise to the normals of the remaining points (30\% perturbation).

Finally, we train a 3D Diffusion Policy \cite{dp3} on the augmented dataset, which demonstrates zero-shot deployment in real-world scenarios (see \Cref{sec:exp:sim2real}).

\section{Experiments}
\label{sec:experiments}
When studying robot learning from videos, we focus on two key dimensions: (i) whether the method can \textbf{scale up}, and (ii) the upper bound of \textbf{task difficulty} it can handle. Therefore, in this section, we systematically evaluate \paper\ along both dimensions.
To be specific, we organize our experiments to answer the following questions:

\textbf{Q1:} What is the usability of the generated data?

\textbf{Q2:} Does \paper\ produce higher-quality dexterous manipulation data compared to existing approaches?

\textbf{Q3:} Can \paper\ handle complex manipulation, including tool-using and long-horizon tasks?

\textbf{Q4:} Does \paper\ enable zero-shot real world deployment?

\begin{table}
\vspace{-5mm}
  \caption{Success rates of 4D object trajectory reconstruction with different methods.}
  \centering
  \begin{tabular}{c|c|c||c|c|c}
    \toprule
    Exp. & Method & Success & Exp. & Method & Success \\
    \midrule
    (a) & TA+RANSAC & 38\% & (d) & DA3+PCR   & 45\% \\
    (b) & TA+PCR    & 11\% &\second (e) &\second  ST2+PCR   &\second 76\% \\
    (c) & VGGT+PCR  & 32\% & \best (f) &\best ST2+FPose &\best 82\% \\
    \bottomrule
  \end{tabular}
  \label{tab:reconstruction}
  \vspace{-2mm}
\end{table}

\subsection{Data Usability}
\label{sec:data_usability}

\paragraph{\textcolor{myOrange}{\textbf{Object Trajectory Estimation}}}
Accurate reconstruction of object trajectories is essential for generating high-quality manipulation data. 
\paper\ accomplishes this through a sequential depth and pose estimation pipeline.
To systematically study this reconstruction framework, we evaluate four depth estimation models: VGGT \cite{vggt}, SpatialTracker v2 \cite{spatialtrackerv2} (ST2), Trace-Anything \cite{traceanything} (TA), and Depth-Anything v3 \cite{da3} (DA3); and three pose estimation methods: RANSAC \cite{ransac}, ColorPCR \cite{colorpcr} (PCR), and FoundationPose++ \cite{FoundationPosePlusPlus} (FPose).

We select 100 short-horizon tasks to evaluate the success rate of these methods.
The results are summarized in \Cref{tab:reconstruction}. 
Experiment (a) uses correspondences estimated by Trace-Anything; 
however, the tracking tends to underestimate object motion, resulting in object trajectories that are substantially smaller than the ground truth. 
Experiments (b)-(e) estimate object poses by first segmenting object point clouds, followed by colored point cloud registration. 
Among these, ST2 produces more temporally consistent depth estimates, yielding a higher success rate.
Finally, experiment (f) combines tracking with 6D pose estimation, achieving the highest reconstruction accuracy.

\paragraph{\textcolor{myOrange}{\textbf{Usability}}}
To comprehensively analyze how video quality influences the usability of generated data, we factorize our evaluation along two orthogonal dimensions: \textit{the quality of input video} and \textit{the difficulty of the target manipulation task}.

\noindent
\textbf{Data Quality Levels.}
We categorize input data into four quality levels:
(1) Sentence-level generation.
Synthesized videos from a text-conditioned video generation model (Wan2.2 \cite{wan2.2} and Veo3 \cite{veo3}).
(2) In-the-wild human videos.
Casually captured or Internet manipulation videos, exhibiting uncontrolled viewpoints, occlusions, and diverse visual artifacts.
(3) Custom-captured videos.
Videos recorded by informed operators with explicit consideration of reconstruction requirements, including minimized occlusion, carefully selected camera viewpoints, and stable object visibility.
(4) Manually corrected videos.
At this level, reconstruction errors in hand and object poses are manually corrected to eliminate estimation failures.

\begin{figure}
\vspace{-5mm}
\centering
\hspace{-7mm}
\includegraphics[width=0.97\linewidth]{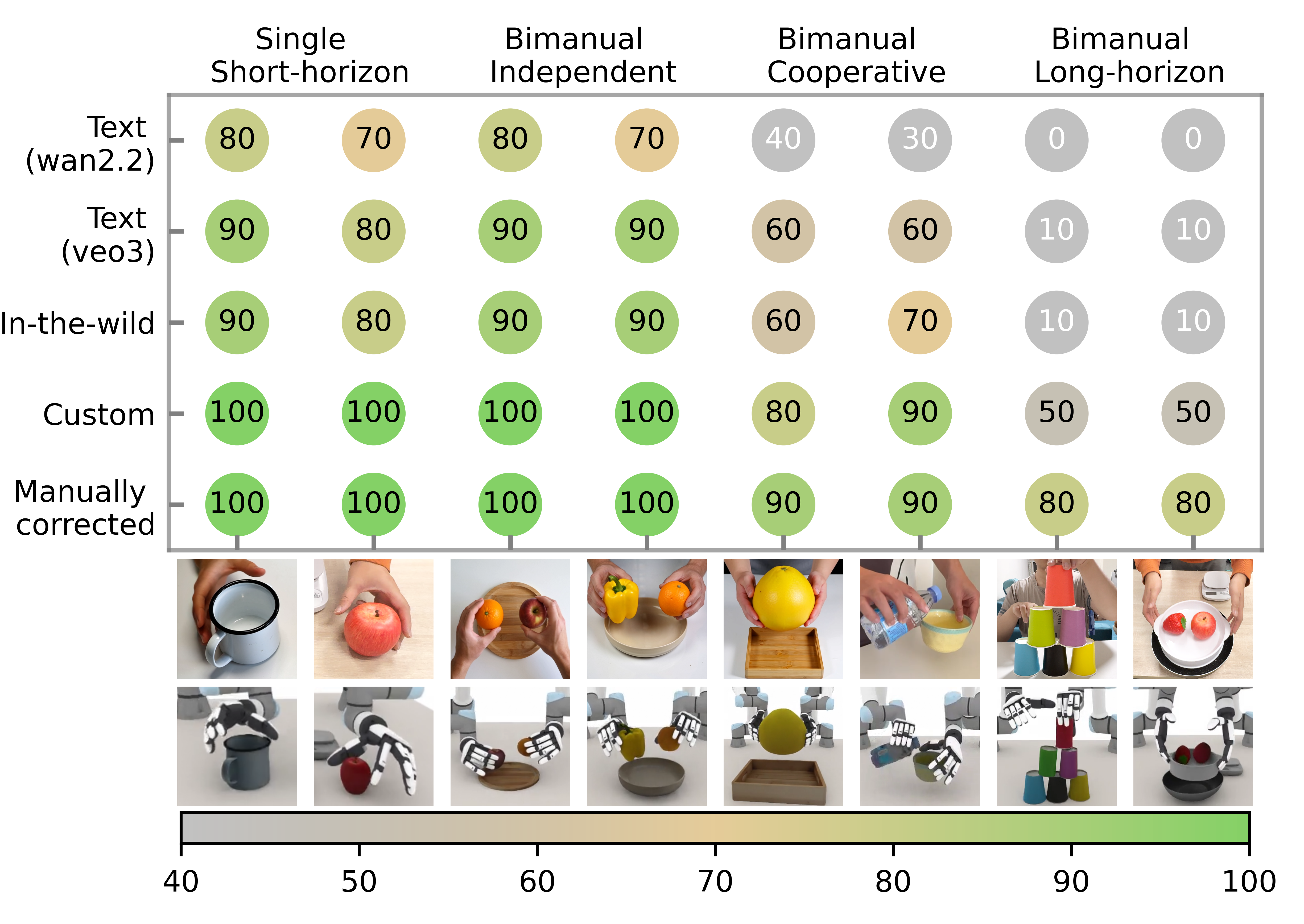}
\captionof{figure}{
Usability evaluation of generated dexterous manipulation data.
The analysis considers two orthogonal factors: \textbf{input data quality} and \textbf{target task difficulty}.
We report data usability rates for two representative manipulation tasks at each difficulty level, with usability visualized on a \textcolor{gray}{gray}-to-\partb{green} color scale.
}
\label{fig:usability}
\vspace{-5mm}
\end{figure}

\noindent
\textbf{Task Difficulty Levels}.
We further stratify manipulation tasks into four categories with increasing complexity:
(1) Unimanual, short-horizon tasks (e.g., grasping, placing).
(2) Cooperative bimanual, short-horizon tasks, 
which require bimanual coordination to manipulate a single object or multiple interacting objects (e.g., holding a pot, pouring water).
(3) Independent bimanual, short-horizon tasks, 
where both hands operate simultaneously but independently (e.g., parallel pouring or grasping).
(4) Cooperative bimanual, long-horizon tasks, 
the most complex setting, involving temporally extended manipulation with sustained coordination across multiple interaction stages (e.g., making a beverage, stacking cups, cutting an apple).

\Cref{fig:usability} illustrates the proportion of usable data (i.e., physically plausible and training-ready samples)
across different data quality and task difficulty levels.
For each difficulty level, we select two representative tasks
and report the average data usability rate of different data sources.
For sentence-level data, we employ Wan2.2~\cite{wan2.2} and Veo3~\cite{veo3}
to generate human videos.
Compared to Wan2.2, Veo3 exhibits stronger temporal consistency
and superior language-following capability.
Consequently, for low-complexity tasks, it achieves a high data usability rate
and remains robust for single-step and two-step manipulations.
Similar to generated videos, casually captured human manipulation videos provide high-quality data for simple tasks.
However, due to factors such as suboptimal camera viewpoints and frequent occlusions,
their reconstruction success rate degrades substantially
as task complexity and temporal horizon increase.
When videos are captured by informed operators,
the reconstruction quality is significantly improved.
This setting yields near-complete data usability for simple tasks
and maintains a considerable success rate on challenging long-horizon tasks.
Finally, with manual correction, long-horizon and fine-grained tasks (e.g., cooking and multi-cup stacking) can be effectively processed, producing training-ready data.

\begin{table}
\vspace{-2mm}
  \centering
  \caption{
  Success rate comparison with baselines across six different task categories. ``-'' in the table indicates failure.
  }
  \resizebox{\columnwidth}{!}{\begin{tabular}{l|ccccccc}
    \toprule
    Task & Put Cup & Grapefruit & Fruits & Pour & Pot & Stack Cups\\
    \midrule
    Rigvid \cite{rigvid} & 96 & - & \textbf{100} & 50 & - & - \\
    Dexman \cite{dexman} & 94  & 98 & - & - & - & - \\
    Ours & \best \textbf{100} &\best \textbf{100} &\best \textbf{100} &\best \textbf{100} &\best \textbf{78} &\best \textbf{52} \\
    \bottomrule
  \end{tabular}}
  \label{tab:sim}
  \vspace{-6mm}
\end{table}

\begin{figure*}
\vspace{-6mm}
\centering
\includegraphics[width=\textwidth]{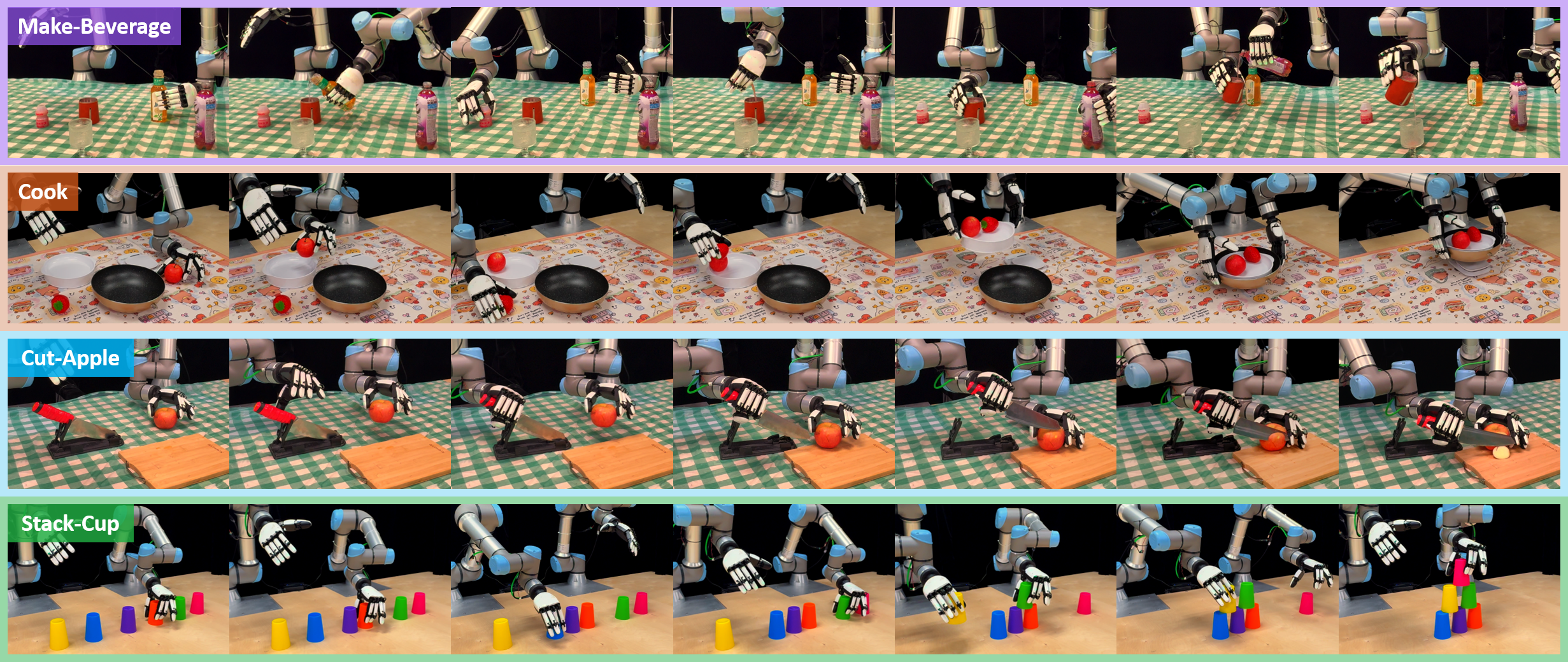}
\caption{\paper\ can generate physically plausible data for long-horizon and fine-grained real-world tasks.}
\label{fig:qualitative}
\vspace{-3mm}
\end{figure*}

\subsection{Data Quality}
\label{sec:exp data quality}

In this section, we evaluate whether the data generated by \paper\ is of sufficient quality for policy learning.
We take a video as input and generate the source trajectory, and then apply data augmentation to produce a dataset of 100 demonstrations.
We then train a DP3 \cite{dp3} policy on the resulting dataset and evaluate its performance in simulation.

\noindent
\textbf{Baseline.}
Only a few prior methods are capable of learning bimanual dexterous manipulation directly from RGB videos. 
We therefore re-implement two recent representative approaches and adapt them for evaluation in our bimanual setting.
RigVid~\cite{rigvid} estimates object poses directly from videos and executes tasks via grasp synthesis and motion planning.
As RigVid is designed for a single-arm gripper, we extend it to a bimanual dexterous hand setting for a fair comparison.
DexMan~\cite{dexman} reproduces actions observed in videos using reinforcement learning.
Since its official codebase has not been released, we re-implement the method following the specifications described in the original paper.

\noindent
\textbf{Tasks.} We select several representative tasks to evaluate data quality:
(1) \textbf{Put Cup}: place a mug onto a plate using one hand.
(2) \textbf{Grapefruit}: lift a grapefruit and place it onto a tray using both hands.
(3) \textbf{Fruits}: pick and place a fruit with each hand.
(4) \textbf{Pour}: hold a bottle with one hand and a bowl with the other to pour liquid.
(5) \textbf{Pot}: place an apple into a pot with one hand, then lift the pot with both hands.
(6) \textbf{Stack Cups}: collaboratively stack six cups into a pyramid using both hands.
\noindent

\begin{figure}
\vspace{-2mm}
\centering
\includegraphics[width=\columnwidth]{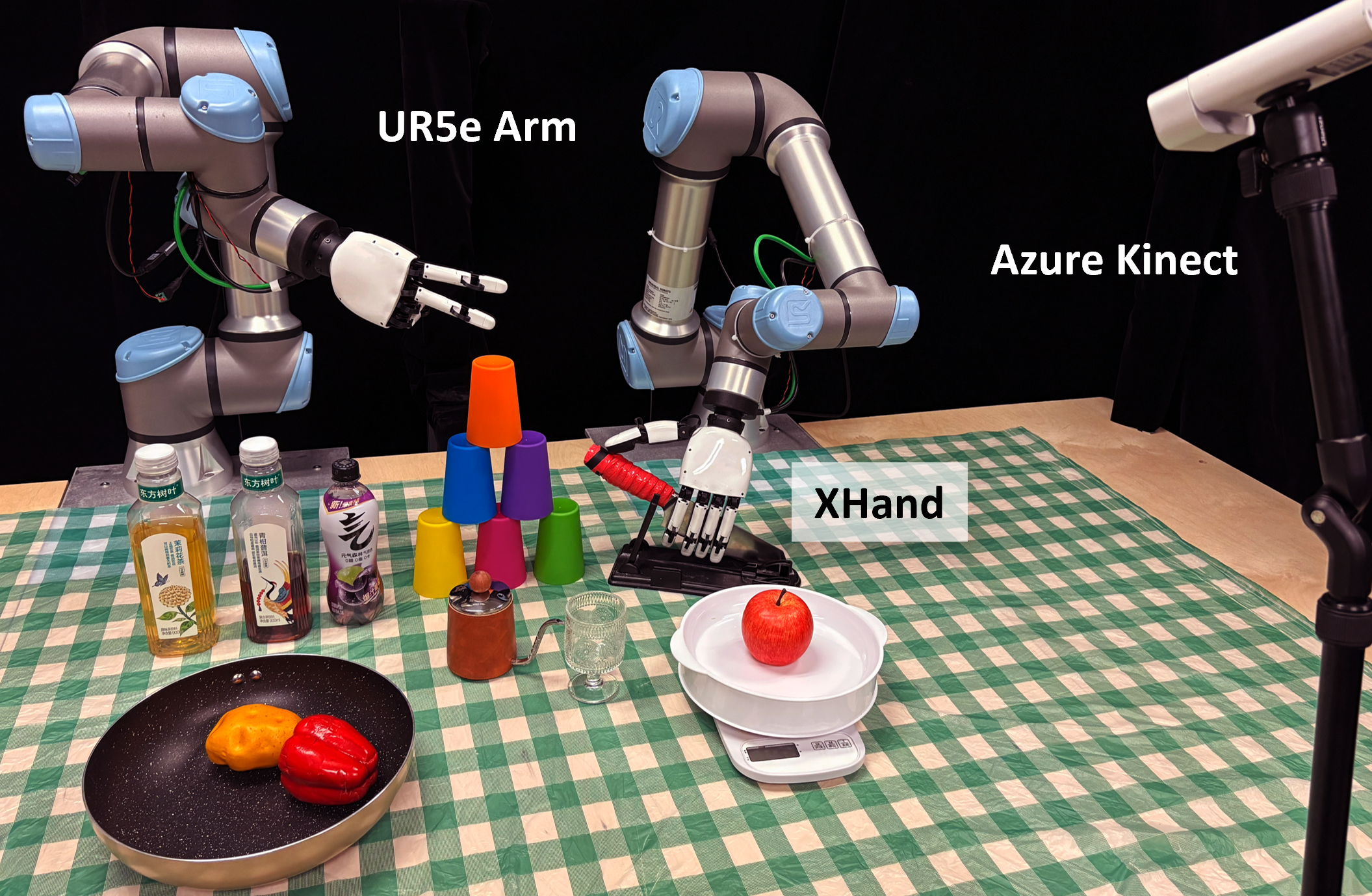}
\caption{Real world experiment setup.}
\label{fig:family}
\vspace{-6mm}
\end{figure}

\noindent
\textbf{Results.}
As shown in \Cref{tab:sim}, \paper\ achieves the highest success rate across all evaluated tasks.
RigVid~\cite{rigvid} performs reasonably well on short-horizon, unimanual tasks, but fails to handle bimanual manipulation and interaction-intensive scenarios.
DexMan~\cite{dexman}, which relies on reinforcement learning to track video-derived trajectories, is highly sensitive to trajectory quality; even small per-frame action inconsistencies can lead to training instability.
Consequently, DexMan succeeds only on short-horizon tasks and struggles with tasks involving multiple sequential actions.
For challenging long-horizon tasks, neither baseline is able to achieve successful execution.
In contrast, \paper\ explicitly decomposes manipulation into subtasks and schedules them over time, while leveraging stable grasp synthesis and keyframe-based motion planning to mitigate the impact of trajectory noise.
As a result, \paper\ achieves near-perfect performance on short-horizon tasks and maintains a high success rate on the long-horizon \textit{Pot} task.
Notably, on the challenging fine-grained \textit{Stack Six Cups} task, \paper\ achieves a success rate of 52\%, demonstrating its strong capability in handling complex dexterous manipulation.

\begin{figure*}
\vspace{-9mm}
\centering
\includegraphics[width=\textwidth]{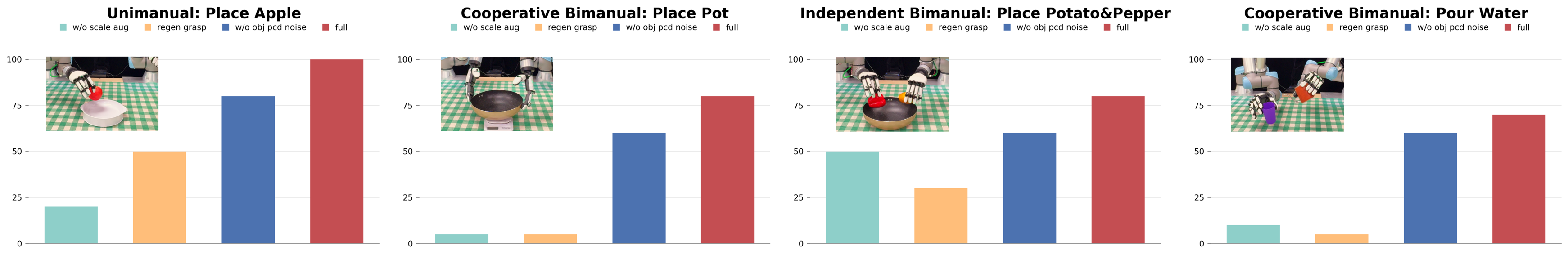}
\caption{
We evaluate \paper\ on four real-world tasks under sim-to-real transfer, including unimanual manipulation, bimanual pick-and-place, and cooperative bimanual interaction tasks.
For each task, we compare three data augmentation ablations.
}
\vspace{-5mm}
\label{fig:sim2real}
\end{figure*}

\begin{figure}
\centering
\includegraphics[width=\columnwidth]{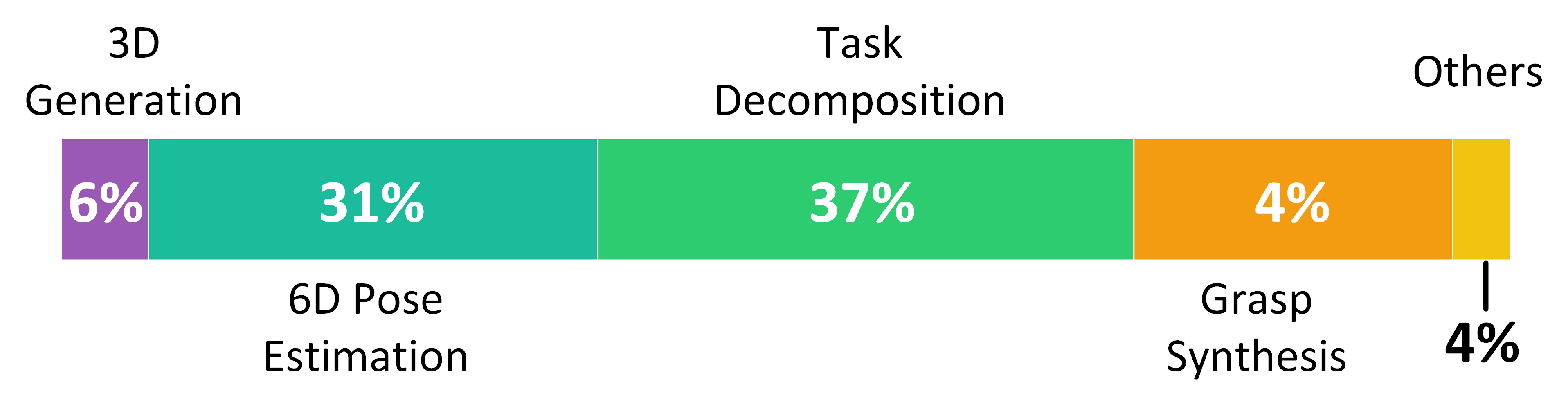}
\vspace{-5mm}
\caption{
Breakdown of error cases. We identify four primary sources of failure; remaining cases are grouped as \emph{Others}.
}
\label{fig:limitation}
\vspace{-5mm}
\end{figure}

\subsection{Task Difficulty}
In \Cref{sec:data_usability}, we study the data usability of \paper\ under varying data quality and task difficulty settings. For short-horizon tasks, \paper\ already demonstrates strong potential for scaling up bimanual dexterous manipulation. In this section, we further evaluate the capabilities of \paper\ from the perspective of task difficulty.
Specifically, we consider four long-horizon tasks:
(1) \textbf{Make-Beverage} is an ultra-long-horizon task, which involves pouring three different beverages into a kettle, shaking to mix them, and finally pouring the mixture into a glass.
(2) \textbf{Cook} is a long-horizon cooperative bimanual task, where apples and strawberries are placed onto a plate, followed by bimanual carrying of the plate and the pot.
(3) \textbf{Cut-Apple} is a tool-using task, requiring a firm grasp of the knife handle and precise cutting of the apple.
(4) \textbf{Stack-Cup} is a long-horizon fine-grained task, which requires accurately stacking six cups into a pyramid.
\Cref{fig:qualitative} presents the qualitative results for these four long-horizon tasks with multi-step hand-object interactions.
\paper\ successfully generates robust action sequences and produces physically plausible data. 
By leveraging force-closure-based grasp synthesis and keyframe-based motion planning, it effectively mitigates compounding errors in the reconstruction process. 
Moreover, the Action-Centric Scheduling Algorithm enables conflict-free scheduling of complex subtasks.

\subsection{Zero-Shot Real World Deployment}
\label{sec:exp:sim2real}

\paper\ reconstructs hand-object interactions from videos at near-metric scale and enables comprehensive data augmentation based on the generated source data, enabling policy learning and zero-shot deployment in the real world.

\noindent
\textbf{Experimental Setup.}
Our real-world setup consists of two UR5e arms equipped with XHands and a Microsoft Azure Kinect depth camera (see \Cref{fig:family}). We evaluate \paper\ on four meta-tasks: (1) Unimanual task: Place Apple
(2) Independent bimanual task: Place Potato\&Pepper
(3) Bimanual grasping task: Place Pot
(4) Cooperative bimanual task: Pour Water.
For each task, we apply data augmentation over object pose and scale, camera pose, and visual observations.

\noindent
\textbf{Ablation Study.}
To quantify the contribution of augmentation to real-world policy success, we conduct three ablation studies:
(1) \textit{w/o scale aug}: scale augmentation is disabled.
(2) \textit{regen grasp}: scale augmentation is applied, but motions are regenerated for each scaled instance instead of being reused.
(3) \textit{w/o obj pcd noise}: visual augmentation is removed.

The detailed experimental results are shown in \Cref{fig:sim2real}.
Across all meta-tasks, \paper\ achieves consistently high zero-shot success rates, demonstrating both the physical plausibility of the reconstructed data and the effectiveness of the proposed data augmentation pipeline.
The ablation results reveal several key insights.
First, removing scale augmentation leads to a significant drop in success rate.
Although the reconstructed trajectories are close to metric scale, dexterous manipulation requires precise spatial perception, and exposing the policy to a distribution of metric-consistent scales during training is crucial for robust real-world execution.
Second, when scale augmentation is applied but grasps and motions are re-generated for each scale, performance degrades drastically, even below the setting without scale augmentation.
This is because motions synthesized at different scales are inconsistent, introducing conflicting supervision signals that hinder imitation learning.
Finally, removing visual augmentations including camera pose and object point cloud noise results in reduced real-world performance.
This can be attributed to inherent noise in point clouds captured by the Kinect depth camera.

\section{Limitations}
\label{sec:limitation}

\noindent
\textbf{Limitation.}
Since \paper\ generates data through the sequential execution of multiple modules, errors may propagate along the pipeline, occasionally rendering the resulting data unusable.
To better understand the sources of failure, we randomly sample 100 failure cases and systematically analyze which module is primarily responsible.
The failure distribution is summarized in \Cref{fig:limitation}.
In addition, we note that the current pipeline cannot handle complex in-hand manipulation.

\noindent
\textbf{Future Work.}
Future work may explore end-to-end data generation approaches to improve both the efficiency and accuracy of data generation.
In addition, adapting the method to support deformable and articulated objects could enhance its practicality, which would require more powerful 3D generation models to handle object deformation and articulated kinematics.
\section{Conclusion}
\label{sec:conclusion}

We present \paper, an automated data generation framework for bimanual dexterous manipulation.
\paper\ directly leverages RGB videos from the Internet or video generation models as input, enabling robot learning without any additional information.
The framework follows a four-stage pipeline: 4D \parta{\textbf{reconstruction}} of hand-object interactions, subtask decomposition and bimanual \partb{\textbf{scheduling}}, \partc{\textbf{action generation}} based on force-closure constraints, and comprehensive data \partd{\textbf{augmentation}}—to produce physically plausible data across a diverse range of manipulation tasks.
Extensive experiments demonstrate \paper’s ability to scale data generation and its effectiveness on challenging tasks with complex hand-object interactions.
These results highlight the framework’s effectiveness in addressing the longstanding challenge of data scarcity in dexterous manipulation.

\bibliographystyle{plainnat}
\bibliography{references}

\newpage

\clearpage
\setcounter{page}{1}
\maketitlesupplementary

\begin{strip}
\centering
\large

\begin{minipage}{0.75\textwidth}

{\Large\bfseries Contents}

\vspace{0.5em}
\hrule height 1pt
\vspace{0.8em}

\tocline{\textbf{A \quad Methodological details}}
        {\pageref{sec:supp method grasp_synthesis}}

\hspace{1.5em}
\tocline{A.1\ Grasp synthesis}
        {\pageref{sec:supp method grasp_synthesis}}

\hspace{1.5em}
\tocline{A.2\ Visual Observation Augmentation}
        {\pageref{sec:supp method augmentation}}

\hspace{1.5em}
\tocline{A.3\ Automatic Data Filter}
        {\pageref{sec:supp method filter}}

\vspace{0.4em}

\tocline{\textbf{B \quad Additional experiments and visualizations}}
        {\pageref{sec:supp exp}}

\hspace{1.5em}
\tocline{B.1\ Simulation Experiments}
        {\pageref{sec:supp exp sim}}

\hspace{1.5em}
\tocline{B.2\ Zero-Shot Real World Deployment}
        {\pageref{sec:supp exp sim2real}}

\hspace{1.5em}
\tocline{B.3\ Runtime Analysis}
        {\pageref{sec:supp exp runtime}}

\vspace{0.4em}

\tocline{\textbf{C \quad Detailed Limitations}}
        {\pageref{sec:supp limitation}}

\vspace{0.8em}
\hrule height 1pt

\end{minipage}
\end{strip}


\section*{A. Method Details}
\label{sec:supp method}

\subsection*{A.1. Grasp Synthesis}
\label{sec:supp method grasp_synthesis}

This section provides additional implementation details of the grasp synthesis module described in the main paper.
While the main text presents the overall formulation and optimization objective, we further detail several components that are essential for practical reproduction.
In particular, we elaborate on (i) how grasp candidates are conditioned on the demonstrated hand-object interaction observed in the input video, (ii) the definition of the distance metric $d(\cdot,\cdot)$ used for grasp ranking, and (iii) the implementation of the grasp stability check procedure used during candidate selection.

\paragraph{\textbf{Conditioning Grasp Candidates on Demonstrated Hand-Object Interaction}}
Given a dexterous hand model described by a URDF, we predefine a set of potential contact points by sampling surface points on the finger links of the hand.
Each sampled contact point is associated with a specific finger and link, and its position and orientation can be computed via forward kinematics given the hand pose.

Conditioned on the input video, we leverage a vision-language model (Qwen3-VL \cite{Qwen3-VL}) to infer the number of fingers actively involved in the demonstrated hand-object interaction.
The predicted finger count $N$ is then used as a structural prior to guide grasp synthesis.
Specifically, we select a subset of contact points corresponding to $N$ distinct fingers from the full contact point set, forming an active contact set $\mathcal{C}_N$.

\begin{figure*}
\centering
\includegraphics[width=\textwidth]{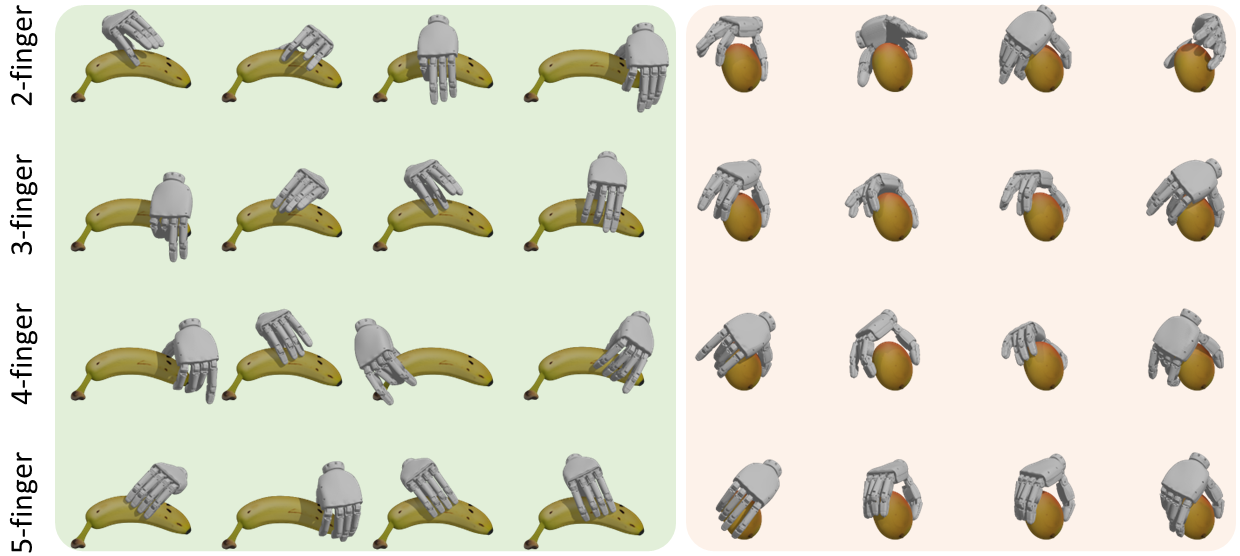}
\vspace{-4mm}
\caption{Visualization of generated grasps.
\paper\ can synthesize diverse grasps for objects with different shapes.
}
\label{fig:grasp}
\end{figure*}

Grasp candidates are generated by solving the grasp optimization problem described in \Cref{method:src generation}
using the restricted contact set $\mathcal{C}_N$.
By conditioning the contact configuration on the inferred finger count, the synthesized grasps explicitly match the demonstrated interaction pattern, thereby enabling the synthesis of $N$-finger grasps that are both physically feasible and consistent with the demonstrated interaction.
\Cref{fig:grasp} visualizes the generated grasps with different numbers of fingers.

\begin{figure}
\vspace{-4mm}
\centering
\includegraphics[width=0.98\linewidth]{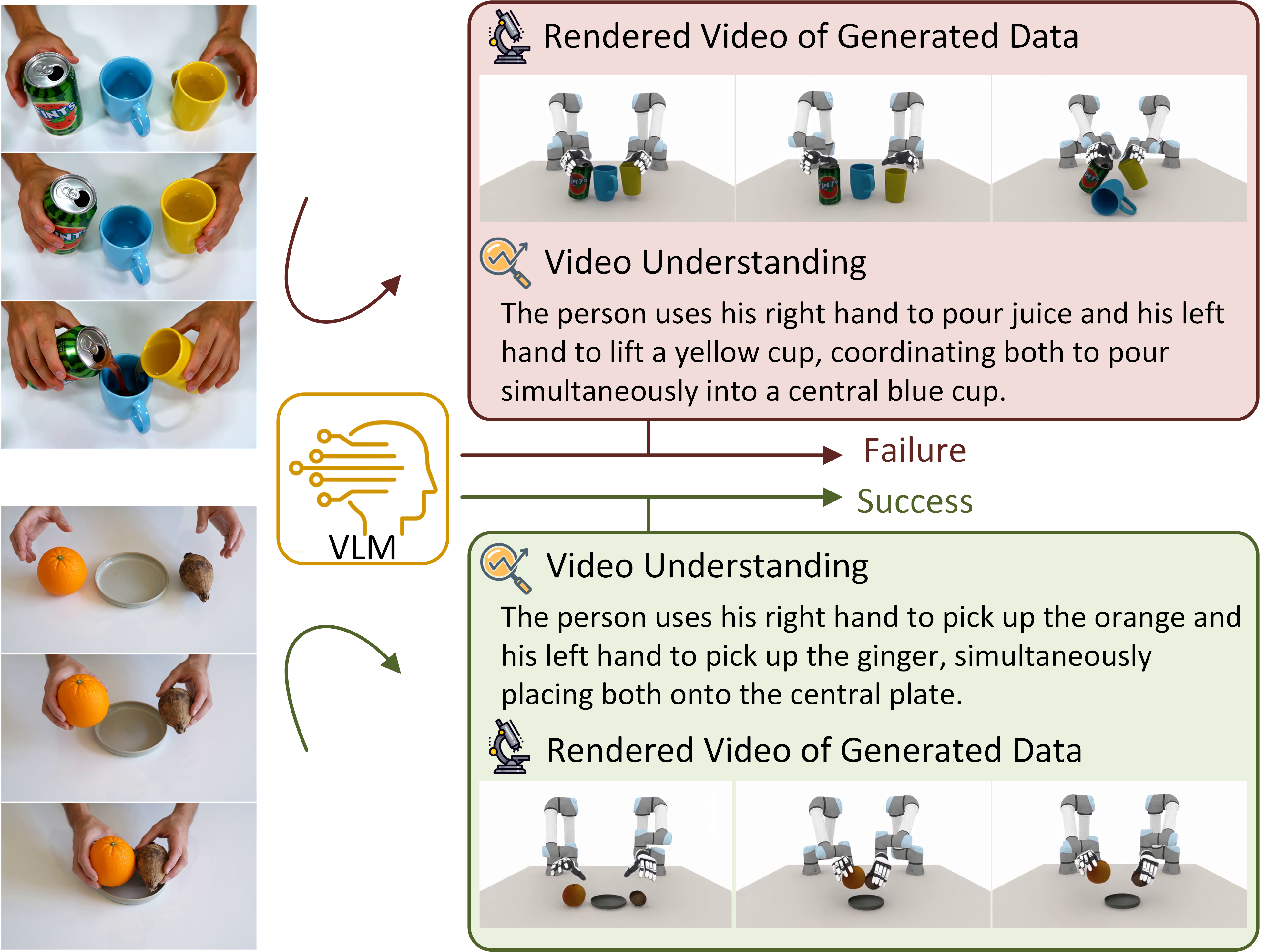}
\caption{
Illustration of automated data filtering. 
We use a Vision-Language Model (VLM) to interpret the human-hand video and generate a textual description, render the synthesized source data, and finally let the VLM verify whether the generated data matches the description.
}
\vspace{-5mm}
\label{fig:filter}
\end{figure}

\paragraph{\textbf{Distance Metric for Grasp Ranking}}
To quantify the consistency between a grasp candidate $g$ and the demonstrated human hand pose $p_{\mathcal{E}_\tau}^t$, we define a distance metric $d(g, p_{\mathcal{E}_\tau}^t)$ that combines translation and rotation errors of the hand.

Specifically, let $(\mathbf{t}_h^g, \mathbf{R}_h^g)$ denote the translation and rotation of hand $h$ in the candidate grasp $g$, and $(\mathbf{t}_h^p, \mathbf{R}_h^p)$ denote the corresponding quantities in the reconstructed human hand pose $p_{\mathcal{E}_\tau}^t$.
We compute the translation error as
\begin{equation}
\Delta \mathbf{t}_h = \mathbf{t}_h^g - \mathbf{t}_h^p, \quad \|\Delta \mathbf{t}_h\|_2,
\end{equation}
and the rotation error as
\begin{equation}
\Delta \mathbf{R}_h = \mathbf{R}_h^g (\mathbf{R}_h^p)^\top, \quad \theta_h = \arccos \Big( \frac{\mathrm{trace}(\Delta \mathbf{R}_h)-1}{2} \Big),
\end{equation}
where $\theta_h$ is the angle of the rotation difference in radians.

Finally, the overall distance metric is computed as a weighted sum over all hands in $\mathcal{E}_\tau$:
\begin{equation}
d(g, p_{\mathcal{E}_\tau}^t) = \sum_{h \in \mathcal{E}_\tau} \lambda_t \|\Delta \mathbf{t}_h\|_2 + \lambda_r \theta_h,
\end{equation}
where $\lambda_t$ and $\lambda_r$ are weights balancing the translation and rotation components.
This metric is used to rank grasp candidates $\mathcal{G}_{o_i}$ in order of their consistency with the demonstrated human behavior.

\paragraph{\textbf{Grasp Stability Evaluation}}
To assess the physical stability of a grasp candidate $g$, we perform a simulation rollout in which the hand(s) move the object from the grasping pose at time $t$ toward the intended target configuration at time $t'$. 

We uniformly sample a point cloud $\mathcal{P}_{o_i} \in \mathbb{R}^{P \times 4}$ from the surface of the object mesh, where each point is represented in homogeneous coordinates. 
During the rollout, we record the resulting object pose $p_{o_i}^{\text{final}}$ and compute the actual relative transformation
\begin{equation}
\mathbf{T}_{\text{sim}} = (p_{o_i}^{t})^{-1} \, p_{o_i}^{\text{final}},
\end{equation}
while the planned relative transformation from the initial to target pose is
\begin{equation}
\mathbf{T}_{o_i}^{t \rightarrow t'} = (p_{o_i}^{t})^{-1} \, p_{o_i}^{t'}.
\end{equation}

Both transformations are applied to the point cloud:
\begin{align}
\mathcal{P}_{\text{sim}} &= \mathcal{P}_{o_i} \, \mathbf{T}_{\text{sim}}^\top, \\
\mathcal{P}_{\text{target}} &= \mathcal{P}_{o_i} \, (\mathbf{T}_{o_i}^{t \rightarrow t'})^\top.
\end{align}

The stability error of the grasp is defined as the mean Euclidean distance between corresponding points:
\begin{equation}
\text{error}(g) = \frac{1}{P} \sum_{p=1}^{P} \left\| \mathcal{P}_{\text{target}}^{(p)} - \mathcal{P}_{\text{sim}}^{(p)} \right\|_2,
\end{equation}
where $P=|\mathcal{P}_{o_i}|$ is the number of sampled points.

A grasp candidate is considered \emph{stable} if $\text{error}(g)$ is below a predefined threshold $\epsilon$. 
We sequentially evaluate the ordered candidates in $\mathcal{G}_{o_i}^{\text{sorted}}$ using this criterion until a stable grasp is found, which is then selected as the final grasp $g^*$.

\begin{figure*}
\centering
\includegraphics[width=\textwidth]{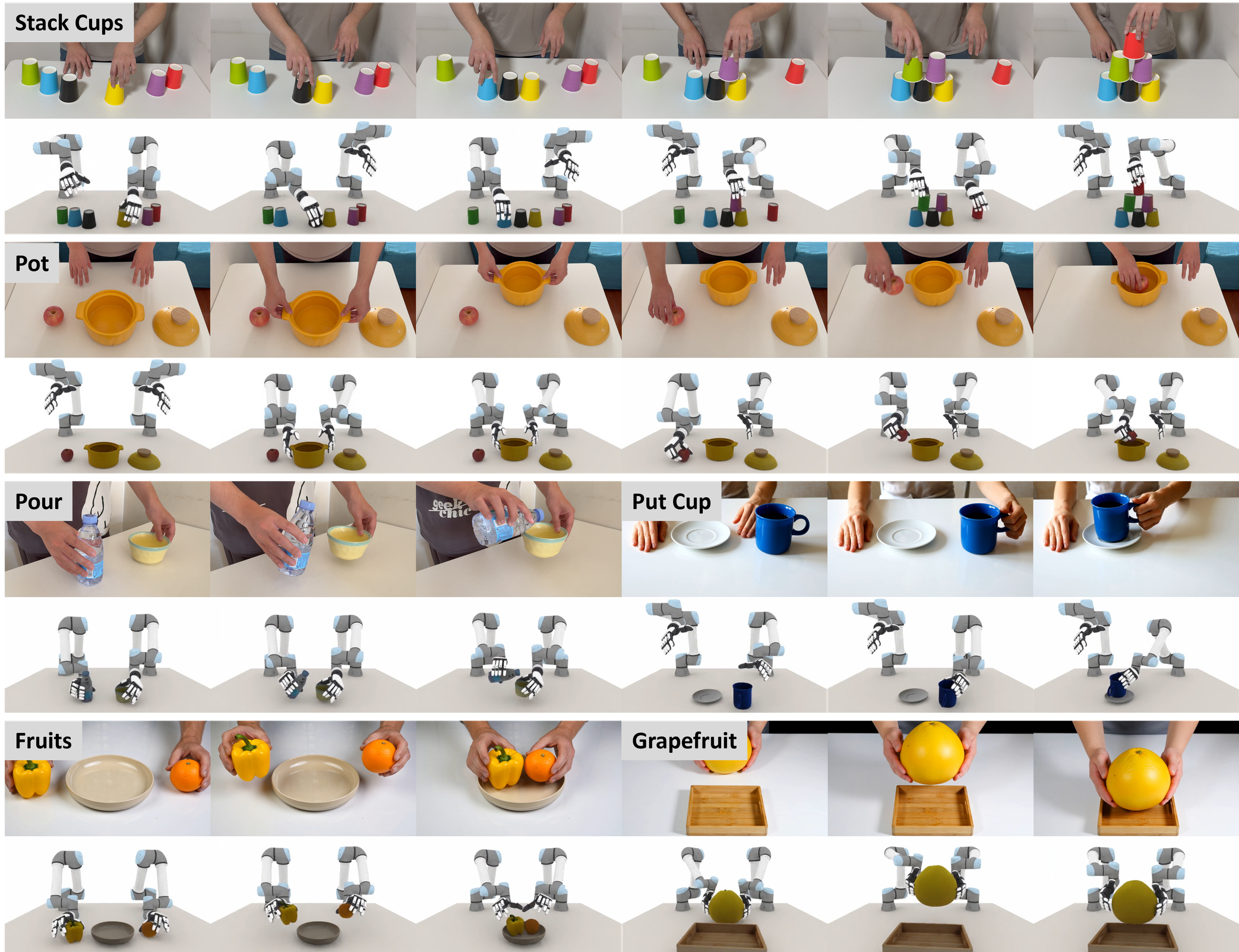}
\caption{
Qualitative results on six simulation tasks.
}
\label{fig:supp_sim2}
\end{figure*}

\subsection*{A.2. Visual Observation Augmentation}
\label{sec:supp method augmentation}

To facilitate sim-to-real transfer and enable zero-shot deployment in real-world scenarios, we apply data augmentation to the visual observations during simulation.
We adopt 3D point clouds as the visual representation, as real-world depth sensors typically exhibit significant noise, missing points, and depth inaccuracies.
To bridge this gap, we explicitly model such imperfections when generating object point clouds in simulation.

Specifically, upon importing an object $o_i$ into the simulator, we uniformly sample a set of surface points from its mesh to construct a reference object point cloud,
\begin{equation}
\mathcal{P}_{o_i}^{\text{ref}} = \{ \mathbf{x}_j \in \mathbb{R}^3 \}_{j=1}^{N},
\end{equation}
where $N$ denotes the number of sampled points.
At each simulation timestep $t$, the object pose $p_{o_i}^t$ is retrieved and converted into a homogeneous transformation matrix $\mathbf{T}_{o_i}^t \in SE(3)$.
The reference point cloud is then transformed to align with the object pose:
\begin{equation}
\mathcal{P}_{o_i}^t = \{ \mathbf{T}_{o_i}^t \, \tilde{\mathbf{x}}_j \}_{j=1}^{N},
\end{equation}
where $\tilde{\mathbf{x}}_j$ denotes the homogeneous coordinates of point $\mathbf{x}_j$.

Let $\mathcal{P}_{\text{scene}}^t = \{ \mathbf{y}_k \}_{k=1}^{M}$ denote the full scene point cloud rendered from the virtual camera at timestep $t$.
To extract object-specific observations, we perform a $k$-nearest neighbor (kNN) search between $\mathcal{P}_{\text{scene}}^t$ and $\mathcal{P}_{o_i}^t$.
A scene point $\mathbf{y}_k$ is classified as an object point if
\begin{equation}
\min_{\mathbf{x} \in \mathcal{P}_{o_i}^t} \| \mathbf{y}_k - \mathbf{x} \|_2 < \delta,
\end{equation}
where $\delta$ is a predefined distance threshold.
All such points are retained to form the extracted object point cloud $\mathcal{P}_{o_i}^{\text{obs}}$.

To simulate sensor noise and partial observations, we further augment the extracted object point cloud $\mathcal{P}_{o_i}^{\text{obs}}$ in two steps.
We first randomly retain a subset of object points,
\begin{equation}
\mathcal{P}_{o_i}^{\text{keep}} \subset \mathcal{P}_{o_i}^{\text{obs}}, \quad
|\mathcal{P}_{o_i}^{\text{keep}}| = 0.85 \, |\mathcal{P}_{o_i}^{\text{obs}}|,
\end{equation}
thereby removing a portion of points to simulate occlusions and missing depth measurements.

Next, we generate an additional set of noisy points to replace the removed observations.
Specifically, we sample a 15\% subset of points from $\mathcal{P}_{o_i}^{\text{keep}}$ with replacement.
For each sampled point $\mathbf{x}_j$, we perturb it along the corresponding surface normal direction:
\begin{equation}
\mathbf{x}_j' = \mathbf{x}_j + \eta \, \mathbf{n}_j, \quad
\eta \sim \mathcal{N}(0, \sigma^2),
\end{equation}
where $\sigma = 0.015$ controls the noise magnitude.
We combine these noisy points with $\mathcal{P}_{o_i}^{\text{keep}}$, resulting in an augmented point cloud with the same cardinality as the original observation.

Finally, the augmented object point cloud is formed by combining the augmented object points and the remaining non-object scene points.
This augmentation procedure produces object observations with realistic sparsity and depth noise characteristics, closely resembling those captured by real-world RGB-D sensors.

Through these augmentations, the policy is exposed to realistic variations in point cloud quality during training, substantially improving robustness and generalization in sim-to-real transfer.

\begin{figure*}
\centering
\includegraphics[width=\textwidth]{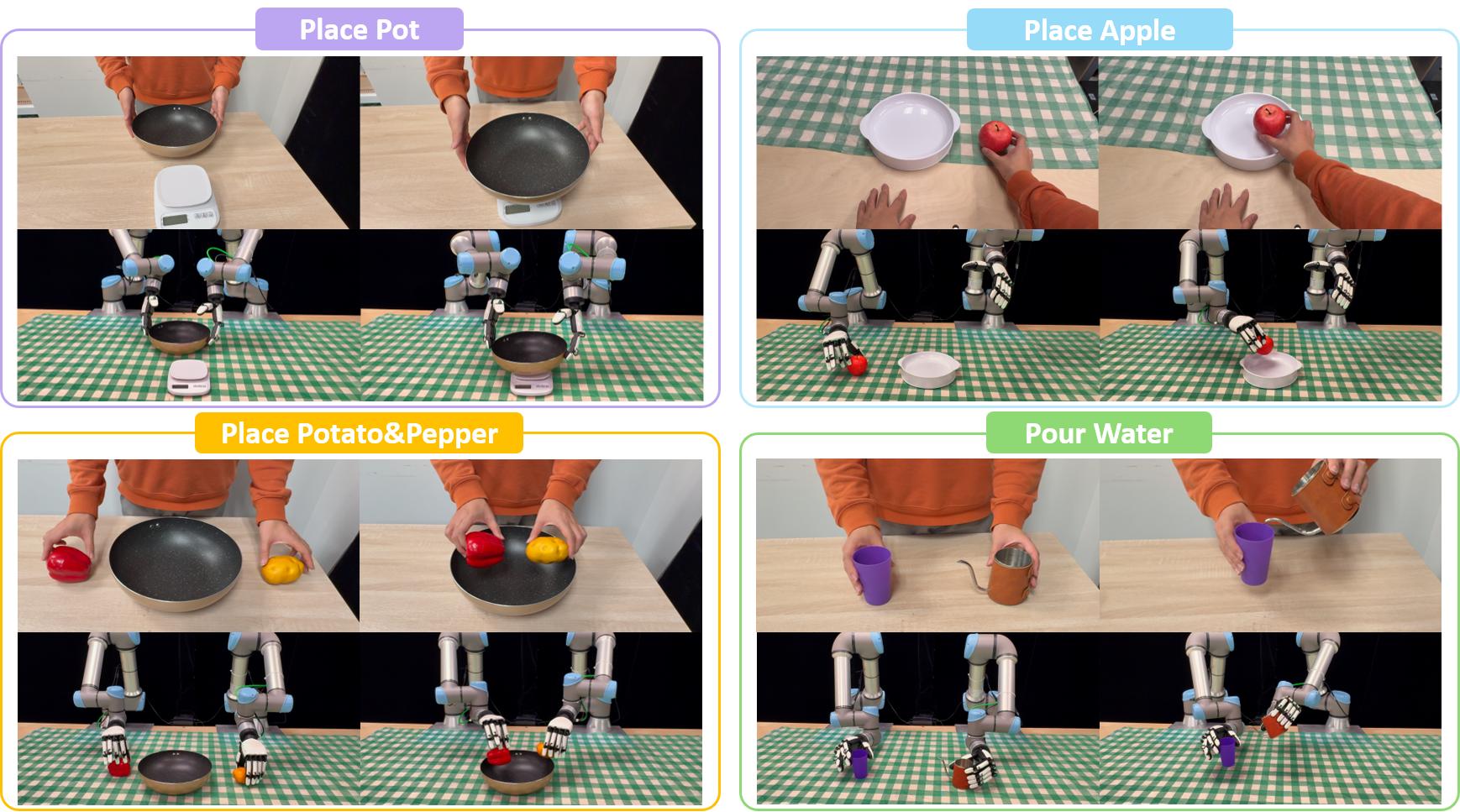}
\caption{
Qualitative results on four real-world sim2real tasks.
}
\label{fig:supp_sim2real}
\end{figure*}

\begin{table*}
  \centering
  \caption{
  Runtime analysis of \paper\ under different input video lengths. All values are measured in seconds.
  }
  \resizebox{\textwidth}{!}{
  \begin{tabular}{c|ccccccccc}
    \toprule
    Video Length & Depth Est. & Hand Pose Est. & Segmentation & 3D Gen. & 6D Pose Est. & Subtask Decomp. & Grasp Synth. & Candidate Sel. & Total \\
    \midrule
    5.0  & 2.7 & 3.4 & 12.9 & 61.9 & 39.1 & 10.6 & 11.3 & 31.2 & 173.1 \\
    10.0 & 5.1 & 8.3 & 22.7 & 62.1 & 41.2 & 21.3 & 11.3 & 29.1 & 201.1 \\
    20.0 & 9.7 & 16.2 & 40.6 & 60.7 & 45.3 & 39.7 & 11.3 & 33.1 & 256.6 \\
    \bottomrule
  \end{tabular}
  }
  \label{tab:runtime}
\end{table*}

\subsection*{A.3. Automatic Data Filter}
\label{sec:supp method filter}

Automated execution is crucial for the \textit{scalability} of data generation. 
While the automated pipeline for synthetic data is described in the main text, obtaining a high-quality dataset also requires automated filtering to remove erroneous samples caused by \textit{compounding errors}.

Our automated data filtering process consists of two steps:  

\begin{enumerate}
    \item During the Subtask Decomposition and Task Scheduling stage, we use Qwen3-VL for video understanding, which produces a predicted video description $t_{\text{pred}}$.
    
    \item After the Source Data Generation stage, we render the synthesized bimanual dexterous manipulation video $V_{\text{synth}}$ using a simulator.
\end{enumerate}

Finally, to determine the usability of the synthesized data, we feed both $t_{\text{pred}}$ and $V_{\text{synth}}$ into Qwen3-VL for evaluation. \Cref{fig:filter} illustrates the VLM-based data filtering process.

\section*{B. Experiments Details}
\label{sec:supp exp}

\subsection*{B.1. Simulation Experiments}
\label{sec:supp exp sim}

In \Cref{sec:exp data quality}, we demonstrate that the data generated by \paper\ meets the requirements for policy training and significantly outperforms existing methods. Here, we present the qualitative results for the evaluated six tasks, as shown in \Cref{fig:supp_sim2}.

\subsection*{B.2. Zero-Shot Real World Deployment}
\label{sec:supp exp sim2real}
In \Cref{sec:exp:sim2real}, we demonstrate that the pipeline of source data generation, data augmentation, and policy learning in \paper\ enables zero-shot real-world deployment. \Cref{fig:supp_sim2real} compares the input human-hand videos with real-robot execution videos for the four meta-tasks. \paper\ effectively transfers skills from human demonstration videos to bimanual dexterous hands, enabling robust performance in complex real-world generalization scenarios.

\subsection*{B.3. Runtime Analysis}
\label{sec:supp exp runtime}

In this section, we provide a detailed analysis of the data generation efficiency of \paper. Specifically, we break down the main sources of computation time, including depth estimation \cite{spatialtrackerv2}, hand pose estimation \cite{wilor}, segmentation \cite{grounded_sam, sam2}, 3D generation \cite{sam3d}, 6D pose estimation \cite{foundpose, FoundationPosePlusPlus}, subtask decomposition \cite{Qwen3-VL}, grasp synthesis, and candidate selection. The overall data generation time is positively correlated with the length of the input video. \Cref{tab:runtime} reports the runtime analysis of \paper\ under different video lengths.
Overall, the results show that the entire data generation time increases with longer input videos, indicating a positive correlation between runtime and video length. Per-frame components such as depth estimation, hand pose estimation, segmentation and subtask decomposition scale with video duration, while downstream modules including 3D generation, grasp synthesis, and candidate selection remain relatively stable, as their computation is dominated by fixed per-video overhead.
Importantly, the overall runtime remains practical for large-scale data generation, with a single video requiring approximately four minutes of processing.

\section*{C. Detailed Limitations}
\label{sec:supp limitation}

In this section, we outline the \textbf{limitations of \paper}:

\begin{itemize}
    \item \textbf{Inability to handle soft or articulated objects:}
    Currently, the 3D generation stage primarily relies on SAM3D~\cite{sam3d}, which assumes rigid object geometry. Consequently, \paper\ cannot yet support manipulation involving articulated or deformable objects. This limitation could potentially be mitigated by integrating more advanced geometry reconstruction or generative modeling methods.
    
    \item \textbf{No support for mobile manipulation:}
    \paper\ is designed for tabletop bimanual manipulation settings. Extending the framework to mobile manipulation scenarios would require explicitly modeling embodiment motion and environmental dynamics.
    
    \item \textbf{Limited performance on long videos without manual intervention:}
    For short-horizon tasks, \paper\ reliably generates accurate bimanual manipulation data. However, because the pipeline executes multiple modules sequentially, errors may accumulate over time. For long input videos, manual intervention is sometimes necessary to maintain accuracy, such as refining VLM-based subtask decomposition or correcting reconstruction artifacts.
    
    \item \textbf{Cannot handle in-hand manipulation:}
    In-hand manipulation is difficult to reconstruct from monocular video due to severe occlusions and limited observability. As a result, \paper\ currently does not include dedicated mechanisms to support such scenarios.

\end{itemize}

\end{document}